\documentclass{article}

\usepackage{template/iclr2027_conference,times}

\usepackage[T1]{fontenc}
\usepackage[utf8]{inputenc}
\usepackage{CJKutf8}
\newcommand{\zh}[1]{\begin{CJK*}{UTF8}{gbsn}#1\end{CJK*}}
\usepackage{microtype}
\usepackage{inconsolata}
\usepackage{latexsym}
\usepackage{graphicx}
\usepackage{algorithm}
\usepackage{algorithmic}
\usepackage{placeins}
\usepackage{amsmath}
\usepackage{booktabs}
\usepackage{array,tabularx}
\usepackage{tikz}
\usetikzlibrary{arrows.meta,positioning,calc}
\usepackage{xcolor}
\usepackage[hidelinks,pdfusetitle]{hyperref}
\usepackage{url}

\newcommand{\authoremail}[1]{\href{mailto:#1}{\normalfont\footnotesize #1}}
\title{HINT-AD: Hierarchical and Interpretable Transformer for Speech-Based Alzheimer's Disease Staging and Marker Discovery}
\author{%
  \begingroup
  \setlength{\tabcolsep}{1.8pt}%
  \begin{tabular}[t]{@{}llll@{}}
    \textbf{Yang Han\textsuperscript{1,2,\textdagger}} &
    \textbf{Jacqueline CK Lam\textsuperscript{2,\textdagger,*}} &
    \textbf{Victor OK Li\textsuperscript{3,2,*}} &
    \textbf{Lawrence YL Cheung\textsuperscript{1,2}}\\
    \authoremail{yanghan001@cuhk.edu.hk} &
    \authoremail{lajacob@elohax.ai} &
    \authoremail{vli@hkust-gz.edu.cn} &
    \authoremail{yllcheung@cuhk.edu.hk}\\
    \authoremail{yhan@elohax.ai} & &
    \authoremail{vli@elohax.ai} &
    \authoremail{yllcheung@elohax.ai}
  \end{tabular}%
  \endgroup\\
  \noalign{\vskip 1.1em}
  \textsuperscript{1}The Chinese University of Hong Kong, Hong Kong SAR, China\\
  \textsuperscript{2}ElohaX Ltd.\\
  \textsuperscript{3}The Hong Kong University of Science and Technology (Guangzhou), China\\[0.2em]
  \textsuperscript{\textdagger}Equal contributions\\
  \textsuperscript{*}Corresponding authors%
}
\hypersetup{pdfauthor={Yang Han, Jacqueline CK Lam, Victor OK Li, Lawrence YL Cheung}}
\iclrfinalcopy

\begin{document}
\maketitle
\lhead{Preprint}
\begin{abstract}
Speech-based studies of cognitive impairment often distinguish patients from cognitively normal controls. We study cross-sectional staging within mild cognitive impairment (MCI): can a picture description distinguish an early MCI (\textit{e}MCI) label from a late MCI (\textit{l}MCI) label at the same assessment? HINT-AD uses marker-specific queries to retrieve acoustic evidence and transcript spans. Four successive stages carry phonological, syntactic, semantic, and cognitive-linguistic evidence forward, so a discourse marker can be interpreted alongside the words and pauses that precede it. We studied 420 English and Mandarin recordings from 218 participants. Because the speech cohorts lack independently adjudicated substages, we transferred labels from the Alzheimer's Disease Neuroimaging Initiative (ADNI) using age, sex, and Mini-Mental State Examination scores. HINT-AD classifies the transferred stage from speech alone. Spanish readings and synthetic Mandarin recordings were used only for training. On real, participant-disjoint holdouts, mean ROC-AUC was 0.77 in each language, compared with 0.67 in English and 0.73 in Mandarin for the strongest speech baselines. For \textit{e}MCI, average precision was 0.65 and 0.67, respectively, and macro F1 was 0.62 in both. Participant-bootstrap 95\% intervals for ROC-AUC were 0.65--0.87 in English and 0.53--0.97 in Mandarin. Masking phonology produced the largest ROC-AUC loss in English, while masking semantics did so in Mandarin. Semantic markers had the strongest positive SHAP attributions in both languages. These findings identify candidate markers of current stage. Their prognostic value requires longitudinal testing against subsequent clinical progression.
\end{abstract}

\section{Introduction}
\label{sec:introduction}
Which speech features distinguish early MCI (\textit{e}MCI) from late MCI (\textit{l}MCI) at a given visit? This is speech-based classification of early versus late MCI. Like a diagnostic assessment, it concerns stage at the current visit. This distinction matters for cohort stratification in Alzheimer's disease (AD) research \citep{edmonds2019early,kasper2020management}. A picture description samples word retrieval, sentence construction, and discourse organization in a single narrative \citep{eyigoz2020linguistic,clarke2021comparison}. We link stage-associated features to the speech that supports them. Prognosis asks whether speech at this visit predicts a later clinical outcome.

These signals span linguistic levels and temporal scales. Pauses before content words can accompany lexical retrieval difficulty; sentence structure reflects grammatical organization; and omissions or inconsistent references affect the narrative as a whole \citep{de2021cognitive,cho2022lexical,gao2024dependency,drummond2015narrative}. Pooled text--audio representations compress these changes into global features, while marker-score vectors summarize linguistic phenomena separately from their speech context. HINT-AD connects marker measurements to their supporting evidence and models interactions across levels.

The prediction target changes the meaning of a benchmark score. \citet{zokaeinikoo2023hierarchical} distinguished probable AD from healthy controls on Pitt. \citet{agbavor2024multilingual} compared MCI with normal cognition, and \citet{pahar2026process2} pooled MCI and dementia against controls. Our comparison lies within MCI and uses the same transferred labels and holdout for every speech model. Hierarchical acoustic encoders organize speech by time scale \citep{chen2023speechformer++,dong2024hafformer}. Here, linguistic markers determine which acoustic or textual evidence enters each stage and remains available to later stages.

We introduce \textbf{HINT-AD}, a \textbf{H}ierarchical and \textbf{IN}terpretable \textbf{T}ransformer for within-MCI staging. Marker-specific queries retrieve acoustic representations or transcript spans, and successive phonological, syntactic, semantic, and cognitive-linguistic stages integrate the resulting summaries. Later stages retain earlier summaries, allowing higher-order representations to use lower-level acoustic-linguistic evidence. Final fusion combines these summaries with transcript and sentence-audio representations. The architecture preserves marker-to-evidence links and accommodates language-specific marker sets within the shared hierarchy.

We evaluate HINT-AD on 420 English and Chinese picture-description recordings from 218 participants, using ADNI-transferred stage labels and a participant-disjoint test set. Training also uses 89 Spanish passage readings and 159 synthetic Mandarin recordings derived from ADNI profiles. None of these 159 ADNI profiles contributed to the stage label transfer of the test samples. Our contributions are:
\begin{itemize}
    \item \textbf{Evidence-linked hierarchical representation.} HINT-AD integrates four linguistic levels through marker-guided Transformer stages, modeling their interactions while retaining the acoustic and textual evidence supporting each summary.
    \item \textbf{Improved multilingual within-MCI staging.} HINT-AD achieves mean ROC-AUC of 0.77 in both languages, compared with 0.67 in English and 0.73 in Mandarin for the strongest speech baselines. It also achieves the highest mean AP and macro F1 in both languages.
    \item \textbf{Interpretable cognitive-linguistic patterns.} Marker-group ablations assess all four linguistic levels, and held-out SHAP analyses identify shared and language-specific marker contributions, with semantic markers providing the strongest positive attributions in both languages.
\end{itemize}

\section{Related Work}
\label{sec:related}
\paragraph{Speech-based cognitive assessment.}
Reviews of speech-based AD assessment describe acoustic, lexical, syntactic, and discourse features, often elicited through picture description \citep{de2020artificial,petti2020systematic,ding2024speech}. Earlier systems used handcrafted linguistic and acoustic features \citep{fraser2016linguistic,ammar2019evaluation}; subsequent models learned text and audio representations \citep{balagopalan2020bert,qin2021exploiting,cai2023exploring,ying2023multimodal}. On PROCESS-2, \citet{pahar2026process2} used linguistic features from picture descriptions to distinguish healthy controls from people with MCI or dementia.

\paragraph{Linguistic markers and hierarchical representations.}
Studies of cognitive decline identify changes in speech timing, phonological fluency, lexical retrieval, and verb processing \citep{de2021cognitive,olmos2023phonological,pekkala2013lexical,williams2021language}. LLM-assisted marker extraction \citep{mo2024leveraging} and DECT \citep{mo2025dect} incorporate fine-grained linguistic knowledge into AD classification. On Pitt, \citet{zokaeinikoo2023hierarchical} used attention over words, sentences, and longitudinal interviews to distinguish probable AD from healthy controls. SpeechFormer++ \citep{chen2023speechformer++} and HAFFormer \citep{dong2024hafformer} organize acoustic computation hierarchically. HINT-AD combines marker-based and hierarchical modeling: marker scores form queries, aligned observations provide evidence, and successive linguistic stages integrate the resulting summaries before classification.

\paragraph{Cross-linguistic modeling.}
DementiaBank \citep{lanzi2023dementiabank} and TAUKADIAL \citep{luz2024connected} support research across speech cohorts and languages. On TAUKADIAL, \citet{agbavor2024multilingual} used language-specific ensembles to detect MCI and predict MMSE, and CogniVoice \citep{cheng2024cognivoice} combined speech and text for the same tasks. Prior work explores cross-language dementia recognition and transfer learning \citep{bertini2022cross,perez2022alzheimer,chen2024cascaded}, as well as LLM-based MCI detection \citep{han2025large}. Closest to our task, \citet{han2025llm} combine generated multilingual transcripts with a contrastive temporal--spatial model spanning several cognitive-status groups. HINT-AD uses recorded speech and its transcripts for within-MCI staging. We evaluate joint English--Chinese learning and Spanish auxiliary training separately, reporting performance for each evaluation language.

\section{HINT-AD}
\label{sec:method}
\subsection{Task and Representation}
\begin{figure}[t]
  \centering
  \includegraphics[width=\columnwidth]{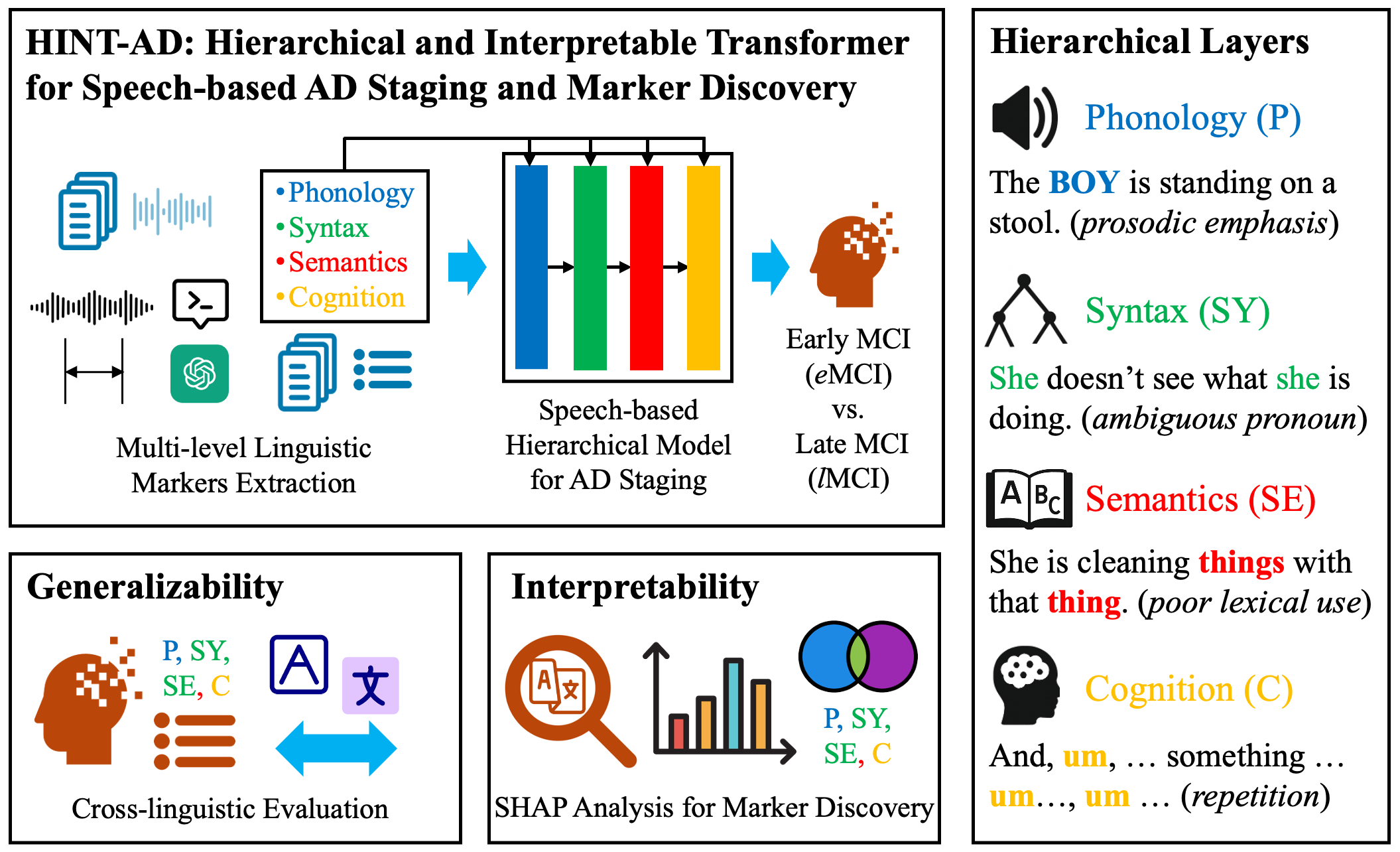}
  \caption{\textbf{Overview of HINT-AD.} Connected speech is represented through transcript and audio encoders together with four levels of interpretable linguistic markers. Marker-specific evidence is integrated progressively from phonology to syntax, semantics, and cognition before final multimodal prediction. The hierarchy is shared across languages, while higher-level marker sets can remain language-specific.}
  \label{fig:architecture}
\end{figure}

Given a connected-speech recording and its automatically generated transcript, HINT-AD predicts the ADNI-transferred \textit{e}MCI or \textit{l}MCI label. Its inputs are transcript tokens $X$, sentence-level audio embeddings $A$, 12 acoustic features $P$, and structured marker measurements with supporting evidence $M$. Four stages integrate phonological, syntactic, semantic, and cognitive-linguistic evidence (Figure~\ref{fig:architecture}); we refer to these domains as phonology, syntax, semantics, and cognition. Each stage passes its marker summaries forward so that subsequent stages can model interactions with earlier evidence.

Multilingual BERT (mBERT) \citep{devlin2019bert} encodes the transcript. Qwen2-Audio \citep{qwen2audio} provides audio embeddings pooled at sentence boundaries identified through forced alignment. Acoustic features measure timing, disfluency, and prosodic variation. Higher-level markers use transcript-level preservation scores and supporting spans to describe syntax, semantics, and cognition. These measurements and evidence links guide the hierarchical representation.

Appendix~\ref{appendix:technical_overview} (Figure~\ref{fig:hint_ad_technical_overview}) details marker queries, evidence retrieval, summary accumulation, and final fusion.

\subsection{Linguistic Markers and Supporting Evidence}
\label{sec:markers}
The marker hierarchy spans acoustic realization, grammatical structure, semantic content, and narrative organization. We adapt the categories of \citet{mo2024leveraging} to English and Chinese within-MCI staging. Marker selection used training transcripts; Appendix~\ref{appendix:hierarchical_layers} describes this development step and lists the definitions.

\paragraph{Phonology.}
Twelve sentence-level features quantify speech rate, pauses, disfluency, emphasis, local repetition, and pitch, energy, and duration variability (Appendix Table~\ref{tab:phonological_markers}). Forced alignment links words to audio intervals; part-of-speech and dependency annotations identify linguistic contexts, such as content words and clause boundaries. The features retain their acoustic measurement interpretation. Appendix~\ref{appendix:speech_examples} (Figure~\ref{fig:speech_examples}) illustrates word timings and pauses in short English and Chinese excerpts from both MCI stages.

\paragraph{Syntax, semantics, and cognition.}
English and Chinese each have five markers per higher-level domain. Syntax describes sentence construction and grammatical relations; semantics describes lexical choice, participant roles, and meaning; cognition describes discourse organization, including omission, repetition, and referential consistency. The shared domains accommodate language-specific phenomena, such as English article and preposition use and Chinese word order (Appendix Table~\ref{tab:syntactic_semantic_cognitive}).

For English and Chinese, an LLM assesses each domain in a separate whole-transcript call. Each marker receives a preservation score in $\{0,0.25,0.5,0.75,1\}$ and up to five supporting spans with line numbers and local scores. Higher values indicate better-preserved function; for omission, they indicate greater content coverage. Unassessable markers receive an explicit flag, a null score, and no evidence. Spanish passage reading permits LLM scoring of nonfunctional repetition only; other higher-level markers remain unassessable, while all 12 acoustic features are retained. HINT-AD learns staging predictions from these annotations together with the speech representations.

Annotations follow a fixed JSON format. We check responses for completeness and validity, regenerate nonconforming responses, and retain accepted annotations for reuse. Three extraction runs are aggregated by within-model consensus. DeepSeek-V4.1-Flash \citep{deepseekai2026deepseekv41flash} supplies the primary markers; GPT-5.6-Luna \citep{openai2026gpt56luna} supplies a second annotation set for the extractor-sensitivity analysis. Appendix~\ref{appendix:prompting_strategy} provides the prompt and response format (Figure~\ref{fig:llm_template}).

\subsection{Marker-Guided Hierarchical Computation}
\label{sec:routing}
HINT-AD assigns two of the first eight mBERT layers to each linguistic stage and uses the final four layers for multimodal fusion. The stages integrate acoustic, grammatical, semantic, and discourse evidence in that order, progressively contextualizing the transcript with marker summaries.

\paragraph{Queries and evidence routing.}
At stage $d$, marker measurements form marker-specific queries $Q_d$. Evidence representations $E_d$ are selected from the current transcript context and projected audio and acoustic features using marker-to-evidence associations. Cross-attention produces marker summaries:
\begin{equation}
    Q_d = \operatorname{EmbedScore}_d(M_d),
    \qquad R_d = \operatorname{CrossAttn}_d(Q_d,E_d),
    \label{eq:routing}
\end{equation}
where $M_d$ contains the measurements and evidence metadata for domain $d$. Phonological queries retrieve aligned acoustic information; higher-level queries retrieve supporting transcript spans and their context. Each summary thus connects a marker measurement to its evidence.

\paragraph{Accumulation across linguistic stages.}
Each stage retains the summaries from earlier stages. With $S_0$ empty, the accumulated representation after stage $d$ is
\begin{equation}
    S_d = [S_{d-1};R_d], \qquad d\in\{1,2,3,4\},
    \label{eq:accumulation}
\end{equation}
where $[\,;\,]$ denotes concatenation. The stage's backbone layers contextualize the transcript with the accumulated summaries, allowing semantic and cognitive-linguistic processing to use earlier acoustic and syntactic evidence. Algorithm~\ref{alg:hint_training} in Appendix~\ref{appendix:training_strategy} gives the forward pass and training loop.

\paragraph{Final fusion and optimization.}
Let $H_4$ be the transcript representation after the four linguistic stages and $U$ the projected sentence-audio and acoustic-feature tokens. Final fusion produces
\begin{equation}
    Z=\operatorname{Fusion}(H_4,U,S_4),
    \qquad \ell=WZ_{\texttt{[CLS]}}+b,
    \label{eq:fusion}
\end{equation}
where $\ell$ contains the two class logits. Classification uses the contextualized original transcript \texttt{[CLS]} token. AdamW minimizes class-weighted cross-entropy, learning how marker summaries and multimodal context jointly support staging.

\subsection{Evidence Links and Marker Attributions}
\label{sec:interpretability}
Evidence-routing attention links each marker summary to the acoustic representations or transcript spans used to form it. SHAP attributions estimate marker contributions to the prediction, and group-masking ablations measure performance changes when an entire linguistic level is withheld. These analyses connect supporting observations with marker-level and domain-level predictive contributions.

We apply grouped KernelSHAP \citep{lundberg2017unified} to the held-out test set using 256 coalitions per recording and fold. Each fold model's raw \textit{e}MCI probability is attributed with transcript words and sentence audio fixed; averaging fold attributions explains the ensemble probability. Reference values are 0.5 for higher-level preservation scores and the language-specific training median for acoustic features. We first average recordings within participants, then summarize participants by language. Positive SHAP values raise the \textit{e}MCI probability relative to the reference; signed means describe direction, and mean absolute values describe magnitude.

\section{Experimental Setup}
\label{sec:experiments}
\subsection{Cohorts, Staging Labels, and Data Split}
\label{sec:data}
The evaluation data comprise MCI picture-description recordings from three cohorts. Pitt \citep{becker1994natural} provides English descriptions, and TAUKADIAL \citep{luz2024connected} provides English and Chinese descriptions; both are accessed through DementiaBank \citep{lanzi2023dementiabank}.\footnote{DementiaBank: \url{https://talkbank.org/dementia/access.html}.} PROCESS-2 \citep{pahar2026process2} contributes its English Cookie Theft Description task.\footnote{PROCESS-2: \url{https://huggingface.co/datasets/CognoSpeak/PROCESS-2}.} Access requires approval from the respective repositories. The Ivanova corpus \citep{ivanova2022discriminating} supplies Spanish passage readings for auxiliary training only.

\paragraph{Transferred staging labels.}
We transfer \textit{e}MCI and \textit{l}MCI labels from the Alzheimer's Disease Neuroimaging Initiative (ADNI) reference cohort, whose staging uses education-adjusted Logical Memory criteria \citep{edmonds2019early}. Separate 5-nearest-neighbor classifiers use standardized age, sex, and Mini-Mental State Examination (MMSE) variables for label transfer. English and Spanish recordings use the ADNI White reference subset; Chinese recordings use the Asian reference subset. HINT-AD predicts these ADNI-transferred stages from speech alone at inference, providing a common staging reference across the speech cohorts.

The English and Chinese data contain 420 picture-description recordings from 218 participants: 137 \textit{e}MCI and 283 \textit{l}MCI recordings. English contributes 288 (104 \textit{e}MCI, 184 \textit{l}MCI), and Chinese contributes 132 (33 \textit{e}MCI, 99 \textit{l}MCI). Cohort counts are 27 for Pitt, 108 for PROCESS-2, and 285 for TAUKADIAL. A fixed participant-disjoint 80--20 split, stratified by stage, cohort, and language, assigns 336 recordings from 175 participants to training and validation and 84 recordings from 43 participants to testing. Separately, 89 Spanish passage readings from 89 participants (19 \textit{e}MCI, 70 \textit{l}MCI) enter auxiliary training only.

\subsection{Speech Preprocessing}
\label{sec:preprocessing}
Qwen3-ASR-1.7B \citep{shi2026qwen3asr} transcribes all three languages. A rule-based procedure removes examiner interventions; we audit removed segments and retained speech for every prepared recording. The Montreal Forced Aligner \citep{mcauliffe2017montreal} supplies word-level audio--text boundaries for duration and pause measurements \citep{rousso2024tradition}.

Stanza \citep{qi2020stanza} supplies English and Spanish part-of-speech and dependency annotations; HanLP \citep{he2021stem} supplies the Chinese annotations. The 12 sentence-level acoustic features use aligned words, inserted pause intervals, linguistic annotations, and z-normalized word-level pitch, energy, and duration. Section~\ref{sec:markers} describes extraction of the eligible higher-level markers.

\subsection{Comparisons and Training Protocol}
\label{sec:evaluation}
\paragraph{Baselines and ablations.}
We evaluate seven baselines: Qwen2-Audio, mBERT, concatenated pooled mBERT and Qwen2-Audio representations (flat fusion), a marker-only multilayer perceptron (MLP), SpeechFormer++, HAFFormer, and CogniVoice \citep{cheng2024cognivoice}. CogniVoice combines Whisper-Tiny, language-specific BERT, and DisVoice acoustic features. All other audio-based models use Qwen2-Audio embeddings, and the hierarchical speech baselines are adapted to binary staging. The marker MLP uses the same four-domain markers without evidence routing. Ablations mask one marker group during both training and evaluation while retaining all backbone layers.

\paragraph{Joint and language-specific training.}
The primary speech models and marker ablations use real English and Mandarin training speech, the 89 Spanish auxiliary readings, and ADNI-derived synthetic Mandarin speech. We generated one recording for each of 159 eligible ADNI profiles: 67 ADNI4 participants had measured English story-recall indicators and 92 ADNI3 participants had imputed indicators. Of the 184 ADNI participants considered for augmentation, we excluded 25 because at least one of their visits was among the five nearest neighbors used to label a test recording or was tied with the fifth neighbor. A synthetic recording entered a fold only if all its Mandarin calibration, transcript-example, and voice donors belonged to that fold's fitting subset. Validation and testing used real recordings only. Appendix~\ref{appendix:computing_infra} describes generation; Table~\ref{tab:joint_language_results} compares training regimes.

\paragraph{Model selection and reporting.}
Five-fold stratified group cross-validation operates within the training partition, with participants disjoint across folds. Joint HINT-AD checkpoints maximize mean English--Chinese validation ROC-AUC, with mean AP breaking ties; language-specific runs use their corresponding metrics. The five selected models predict the same fixed held-out test set. We compute ROC-AUC, average precision (AP), and macro F1 for each model and report the mean and standard deviation (SD) across the five fits, separately by language. Macro F1 gives both stages equal weight. For each model configuration and language, the decision threshold maximizes macro F1 on pooled out-of-fold training predictions. For primary HINT-AD, we also report percentile 95\% confidence intervals (CIs) from 5,000 participant-bootstrap draws per language. Within each stage we resampled participants with replacement, kept their recordings together, scored each fit at the fixed language-specific cutoff for macro F1, and averaged each metric across fits.

AP complements ROC-AUC under class imbalance \citep{saito2015precision}. With \textit{e}MCI as the positive class, AP weights precision at each observed threshold by the gain in recall. With few positive recordings, recall changes in large steps, and linear interpolation between precision--recall points can overestimate area \citep{davis2006relationship}. We pool 57 English test recordings across the three cohorts, which share a picture-description task and a staging target. The Chinese recordings come from TAUKADIAL. Appendix Table~\ref{tab:cohort_results} reports each cohort separately.

HINT-AD uses a learning rate of $2\times10^{-5}$, batch size 8, and at most 10 epochs, with 10\% linear warmup and early-stopping patience 3. Appendix~\ref{appendix:computing_infra} reports the remaining settings and computing environment.

\section{Experimental Results}
\label{sec:results}
\subsection{Held-Out Within-MCI Discrimination}
\label{sec:performance}
HINT-AD achieves the highest mean ROC-AUC, AP, and macro F1 among the evaluated speech-based models in both languages (Table~\ref{tab:performance_results}). In English, HINT-AD exceeded the strongest speech baseline by 0.10 in ROC-AUC, 0.10 in AP, and 0.06 in macro F1. In Mandarin, the corresponding margins were 0.04, 0.01, and 0.05. Mean ROC-AUC was 0.77 in both languages. Table~\ref{tab:performance_results} reports the mean and standard deviation of each metric across the five model fits. Table~\ref{tab:participant_uncertainty} (Appendix~\ref{appendix:computing_infra}) shows that the participant-bootstrap 95\% CIs for all three HINT-AD metrics were wider in Mandarin than in English, consistent with the smaller Mandarin test cohort. Appendix Table~\ref{tab:cohort_results} gives dataset-specific results. Logistic regression using the three label-transfer inputs (age, sex, and MMSE) achieved ROC-AUC values of 0.91 in English and 0.83 in Mandarin (Appendix Table~\ref{tab:metadata_results}). HINT-AD, by contrast, predicts stage from speech alone and does not require an MMSE score at inference.

\begin{table}[htbp]
\centering
\caption{Held-out performance with Spanish auxiliary and ADNI-derived Mandarin training speech. Values are mean $\pm$ SD across five fits. Bold and underline denote the highest and second-highest displayed means, including ties. Flat fusion combines mBERT and Qwen2-Audio; primary markers are from DeepSeek-V4.1-Flash.}
\label{tab:performance_results}
\footnotesize
\setlength{\tabcolsep}{2pt}
\begin{tabular*}{\linewidth}{@{\extracolsep{\fill}}lcccccc@{}}
\toprule
 & \multicolumn{3}{c}{\textbf{English}} & \multicolumn{3}{c}{\textbf{Chinese}} \\
\cmidrule(lr){2-4}\cmidrule(l){5-7}
\textbf{Model} & \textbf{ROC-AUC} & \textbf{AP} & \textbf{Macro F1} & \textbf{ROC-AUC} & \textbf{AP} & \textbf{Macro F1} \\
\midrule
Qwen2-Audio & 0.64\,\ensuremath{\pm}\,0.11 & \underline{0.55}\,\ensuremath{\pm}\,0.10 & 0.53\,\ensuremath{\pm}\,0.12 & 0.63\,\ensuremath{\pm}\,0.10 & 0.50\,\ensuremath{\pm}\,0.08 & 0.44\,\ensuremath{\pm}\,0.09 \\
mBERT & 0.58\,\ensuremath{\pm}\,0.08 & 0.46\,\ensuremath{\pm}\,0.06 & 0.44\,\ensuremath{\pm}\,0.12 & 0.65\,\ensuremath{\pm}\,0.07 & 0.59\,\ensuremath{\pm}\,0.09 & \underline{0.57}\,\ensuremath{\pm}\,0.12 \\
Flat fusion & \underline{0.67}\,\ensuremath{\pm}\,0.04 & 0.52\,\ensuremath{\pm}\,0.04 & 0.51\,\ensuremath{\pm}\,0.11 & 0.70\,\ensuremath{\pm}\,0.05 & 0.65\,\ensuremath{\pm}\,0.05 & 0.53\,\ensuremath{\pm}\,0.09 \\
Marker MLP & 0.62\,\ensuremath{\pm}\,0.07 & 0.45\,\ensuremath{\pm}\,0.06 & 0.45\,\ensuremath{\pm}\,0.09 & 0.72\,\ensuremath{\pm}\,0.14 & \underline{0.66}\,\ensuremath{\pm}\,0.18 & 0.47\,\ensuremath{\pm}\,0.15 \\
SpeechFormer++ & 0.61\,\ensuremath{\pm}\,0.08 & 0.45\,\ensuremath{\pm}\,0.08 & 0.38\,\ensuremath{\pm}\,0.13 & 0.62\,\ensuremath{\pm}\,0.13 & 0.58\,\ensuremath{\pm}\,0.13 & 0.39\,\ensuremath{\pm}\,0.14 \\
HAFFormer & 0.61\,\ensuremath{\pm}\,0.07 & 0.53\,\ensuremath{\pm}\,0.08 & 0.43\,\ensuremath{\pm}\,0.03 & \underline{0.73}\,\ensuremath{\pm}\,0.07 & 0.65\,\ensuremath{\pm}\,0.08 & 0.43\,\ensuremath{\pm}\,0.12 \\
CogniVoice & \underline{0.67}\,\ensuremath{\pm}\,0.03 & 0.53\,\ensuremath{\pm}\,0.07 & \underline{0.56}\,\ensuremath{\pm}\,0.09 & 0.65\,\ensuremath{\pm}\,0.08 & 0.50\,\ensuremath{\pm}\,0.14 & 0.46\,\ensuremath{\pm}\,0.08 \\
\textbf{HINT-AD} & \textbf{0.77\,\ensuremath{\pm}\,0.03} & \textbf{0.65\,\ensuremath{\pm}\,0.04} & \textbf{0.62\,\ensuremath{\pm}\,0.03} & \textbf{0.77\,\ensuremath{\pm}\,0.10} & \textbf{0.67\,\ensuremath{\pm}\,0.11} & \textbf{0.62\,\ensuremath{\pm}\,0.10} \\
\bottomrule
\end{tabular*}
\end{table}

\subsection{Effects of Joint and Auxiliary Training}
\label{sec:joint}
Table~\ref{tab:joint_language_results} compares four training regimes on the same real test recordings. Adding ADNI-derived Mandarin speech to joint training with Spanish readings increased mean ROC-AUC from 0.74 to 0.77 in English and from 0.76 to 0.77 in Mandarin. Mean AP rose from 0.62 to 0.65 and from 0.65 to 0.67, respectively.

\begin{table}[htbp]
\centering
\caption{HINT-AD performance by training regime. Language-only models are fit separately. ADNI denotes synthetic Mandarin training speech. Values are mean $\pm$ SD across five fold-trained models.}
\label{tab:joint_language_results}
\footnotesize
\setlength{\tabcolsep}{2pt}
\begin{tabular*}{\linewidth}{@{\extracolsep{\fill}}lcccccc@{}}
\toprule
 & \multicolumn{3}{c}{\textbf{English}} & \multicolumn{3}{c}{\textbf{Chinese}} \\
\cmidrule(lr){2-4}\cmidrule(l){5-7}
\textbf{Training} & \textbf{ROC-AUC} & \textbf{AP} & \textbf{Macro F1} & \textbf{ROC-AUC} & \textbf{AP} & \textbf{Macro F1} \\
\midrule
Language only & 0.62\,\ensuremath{\pm}\,0.07 & 0.49\,\ensuremath{\pm}\,0.07 & 0.38\,\ensuremath{\pm}\,0.12 & 0.67\,\ensuremath{\pm}\,0.12 & 0.54\,\ensuremath{\pm}\,0.07 & 0.48\,\ensuremath{\pm}\,0.11 \\
Joint English--Mandarin & 0.68\,\ensuremath{\pm}\,0.08 & 0.56\,\ensuremath{\pm}\,0.10 & 0.61\,\ensuremath{\pm}\,0.07 & 0.68\,\ensuremath{\pm}\,0.09 & 0.55\,\ensuremath{\pm}\,0.07 & 0.55\,\ensuremath{\pm}\,0.10 \\
Joint + Spanish & 0.74\,\ensuremath{\pm}\,0.03 & 0.62\,\ensuremath{\pm}\,0.03 & \textbf{0.62\,\ensuremath{\pm}\,0.07} & 0.76\,\ensuremath{\pm}\,0.11 & 0.65\,\ensuremath{\pm}\,0.14 & \textbf{0.66\,\ensuremath{\pm}\,0.14} \\
Joint + Spanish + ADNI & \textbf{0.77\,\ensuremath{\pm}\,0.03} & \textbf{0.65\,\ensuremath{\pm}\,0.04} & \textbf{0.62\,\ensuremath{\pm}\,0.03} & \textbf{0.77\,\ensuremath{\pm}\,0.10} & \textbf{0.67\,\ensuremath{\pm}\,0.11} & 0.62\,\ensuremath{\pm}\,0.10 \\
\bottomrule
\end{tabular*}
\end{table}

\subsection{Contributions of the Four Linguistic Levels}
\label{sec:ablations}
Masking any marker group lowered mean ROC-AUC, AP, and macro F1 in both languages (Table~\ref{tab:ablation_results}). Phonology produced the largest ROC-AUC loss in English (0.77 to 0.73); semantics produced the largest loss in Mandarin (0.77 to 0.68). Masking cognition produced the lowest Mandarin macro F1 (0.45).

\begin{table}[htbp]
\centering
\caption{Marker-group ablations with the full, ADNI-augmented training configuration. Values are mean $\pm$ SD across five fold-trained models. Each group is masked during training and evaluation, with all backbone layers retained.}
\label{tab:ablation_results}
\footnotesize
\setlength{\tabcolsep}{2pt}
\begin{tabular*}{\linewidth}{@{\extracolsep{\fill}}lcccccc@{}}
\toprule
 & \multicolumn{3}{c}{\textbf{English}} & \multicolumn{3}{c}{\textbf{Chinese}} \\
\cmidrule(lr){2-4}\cmidrule(l){5-7}
\textbf{Masked group} & \textbf{ROC-AUC} & \textbf{AP} & \textbf{Macro F1} & \textbf{ROC-AUC} & \textbf{AP} & \textbf{Macro F1} \\
\midrule
\textbf{None (full model)} & \textbf{0.77\,\ensuremath{\pm}\,0.03} & \textbf{0.65\,\ensuremath{\pm}\,0.04} & \textbf{0.62\,\ensuremath{\pm}\,0.03} & \textbf{0.77\,\ensuremath{\pm}\,0.10} & \textbf{0.67\,\ensuremath{\pm}\,0.11} & \textbf{0.62\,\ensuremath{\pm}\,0.10} \\
Phonology & 0.73\,\ensuremath{\pm}\,0.06 & 0.62\,\ensuremath{\pm}\,0.07 & 0.56\,\ensuremath{\pm}\,0.12 & 0.72\,\ensuremath{\pm}\,0.09 & 0.59\,\ensuremath{\pm}\,0.16 & 0.54\,\ensuremath{\pm}\,0.12 \\
Syntax & 0.75\,\ensuremath{\pm}\,0.05 & 0.62\,\ensuremath{\pm}\,0.06 & 0.60\,\ensuremath{\pm}\,0.10 & 0.72\,\ensuremath{\pm}\,0.13 & 0.61\,\ensuremath{\pm}\,0.11 & 0.51\,\ensuremath{\pm}\,0.10 \\
Semantics & 0.74\,\ensuremath{\pm}\,0.04 & 0.61\,\ensuremath{\pm}\,0.04 & 0.53\,\ensuremath{\pm}\,0.09 & 0.68\,\ensuremath{\pm}\,0.18 & 0.60\,\ensuremath{\pm}\,0.14 & 0.52\,\ensuremath{\pm}\,0.06 \\
Cognition & 0.75\,\ensuremath{\pm}\,0.06 & 0.63\,\ensuremath{\pm}\,0.06 & 0.57\,\ensuremath{\pm}\,0.10 & 0.74\,\ensuremath{\pm}\,0.11 & 0.60\,\ensuremath{\pm}\,0.15 & 0.45\,\ensuremath{\pm}\,0.08 \\
\bottomrule
\end{tabular*}
\end{table}

\subsection{Marker Contributions to Held-Out Predictions}
\label{sec:shap}
Table~\ref{tab:marker_shap} ranks the participant-balanced signed mean contributions to the raw \textit{e}MCI probability. Semantic markers account for all five leading positive contributions in English and four of five in Mandarin. Role identification ranks first in English, followed by lexical use and scene integration; thematic coherence ranks first in Mandarin, followed by role identification and verb use.

\begin{table}[t]
\centering
\caption{Five largest positive and negative participant-balanced signed mean SHAP contributions per language on the held-out test set. Positive values shift raw \textit{e}MCI probability upward; negative values shift it downward. Values are on the 0--1 probability scale. Attributions are averaged across the five fold models to explain their mean predicted \textit{e}MCI probability for each recording.}
\label{tab:marker_shap}
\small
\setlength{\tabcolsep}{5pt}
\begin{tabular}{@{}cllr@{}}
\toprule
\textbf{Rank} & \textbf{Level} & \textbf{Marker} & \textbf{Mean SHAP} \\
\midrule
\multicolumn{4}{@{}l}{\textbf{English: toward \textit{e}MCI}} \\
1 & Semantics & Role identification & $+0.01038$ \\
2 & Semantics & Lexical use & $+0.00855$ \\
3 & Semantics & Scene integration & $+0.00851$ \\
4 & Semantics & Contrast use & $+0.00416$ \\
5 & Semantics & Synonym use & $+0.00171$ \\
\addlinespace
\multicolumn{4}{@{}l}{\textbf{English: toward \textit{l}MCI}} \\
1 & Phonology & Mean pause duration within clause & $-0.00069$ \\
2 & Phonology & Word-duration variability & $-0.00059$ \\
3 & Phonology & Clause-final pause duration & $-0.00052$ \\
4 & Phonology & Energy variability & $-0.00039$ \\
5 & Phonology & Pitch variability & $-0.00022$ \\
\addlinespace
\multicolumn{4}{@{}l}{\textbf{Chinese: toward \textit{e}MCI}} \\
1 & Semantics & Thematic coherence & $+0.00228$ \\
2 & Semantics & Role identification & $+0.00177$ \\
3 & Semantics & Verb use & $+0.00169$ \\
4 & Semantics & Lexical use & $+0.00144$ \\
5 & Phonology & Energy variability & $+0.00030$ \\
\addlinespace
\multicolumn{4}{@{}l}{\textbf{Chinese: toward \textit{l}MCI}} \\
1 & Phonology & Pause ratio & $-0.00041$ \\
2 & Phonology & Disfluency count & $-0.00025$ \\
3 & Semantics & Synonym use & $-0.00017$ \\
4 & Phonology & Mean pause before content words & $-0.00012$ \\
5 & Phonology & Clause-final pause duration & $-0.00010$ \\
\bottomrule
\end{tabular}
\end{table}

The largest negative contributions were mean within-clause pause duration in English and pause ratio in Mandarin. Clause-final pause duration had a negative mean contribution in both languages. These signs describe shifts in the fitted model relative to its reference values, not differences between the two stage groups.

We also examined whether the ten speech markers identified by SHAP in each language differed between early and late MCI. Using two-sided Welch's $t$-tests to compare participant-level mean marker values, we found differences for four markers in English and three in Mandarin ($p<0.05$; Appendix Table~\ref{tab:marker_stage}). Lexical use was the only selected marker that differed in both languages: the \textit{e}MCI group scored higher than the \textit{l}MCI group for specific, context-appropriate word choice.

\subsection{Marker Repeatability and Extractor Sensitivity}
\label{sec:consistency}
For English and Chinese markers, DeepSeek-V4.1-Flash achieves 90.6\% three-run agreement, 92.3--96.6\% pairwise agreement, and Cohen's $\kappa$ of 0.885--0.949. GPT-5.6-Luna achieves 81.8\% three-run agreement, 87.4--88.3\% pairwise agreement, and $\kappa$ of 0.825--0.837. Spanish follows the separate sparse marker policy defined for passage reading and contributes through auxiliary training.

Repetition and omission scores from both extractors also show the expected associations with transcript-derived repetition and content coverage in English and Chinese (Appendix~\ref{appendix:marker_consistency}, Table~\ref{tab:marker_consistency}). Together with repeated-run agreement, these results provide complementary automated evidence for consistency of the repetition and omission annotations across both languages.

With the same augmented training data, DeepSeek markers yielded higher mean ROC-AUC and AP in both languages than GPT markers (Appendix Table~\ref{tab:cross_llm_results}). GPT markers yielded higher English macro F1 (0.65 vs.\ 0.62); DeepSeek markers yielded higher Mandarin macro F1 (0.62 vs.\ 0.55).

\subsection{Interpretation of Speech Markers and Model Behavior}
\label{sec:discussion}
\textbf{Semantic content and lexical retrieval.}
Role identification and lexical use contribute positively to the \textit{e}MCI probability in both languages, linking the shared signal to how speakers identify scene participants and express relevant concepts. In longitudinal picture descriptions from autopsy-confirmed AD cases, \citet{ahmed2013connected} found that semantic and lexical content tracked linguistic decline from MCI through later disease stages. The Chinese verb-use contribution has related support from \citet{lai2013action}: action fluency and action naming were associated with dementia severity in Chinese-speaking participants. These findings provide a linguistic basis for interpreting the semantic contributions in HINT-AD.

\textbf{Narrative organization and reference.}
English scene integration and Chinese thematic coherence capture how details form a connected account. \citet{drummond2015narrative} found incomplete narratives and irrelevant propositions in amnestic MCI, with broader coherence and referential difficulties in AD. In our excerpts, English scene integration captures whether the speaker connects an action to its consequences; Mandarin thematic coherence captures whether the account maintains a common thread across events.

\textbf{Speech timing and fluency.}
Within-clause pause duration in English and the proportion of time spent in pauses in Mandarin had the strongest mean attributions toward \textit{l}MCI among the selected markers. \citet{de2021cognitive} linked conversational timing to lexico-semantic and executive performance in MCI and AD, connecting temporal organization with the demands of language production.

\textbf{Implications for hierarchical modeling and marker discovery.}
The complementary contributions of meaning, discourse, and timing are consistent with the semantic, syntactic, acoustic, and information-content factors identified in DementiaBank speech by \citet{fraser2016linguistic}. Semantics had the strongest positive marker contributions in both languages, but the largest ROC-AUC loss after masking an entire level occurred for phonology in English and semantics in Mandarin. SHAP describes shifts in a fitted model; masking and retraining tests what the remaining levels can recover.

\section{Conclusion}
\label{sec:conclusion}
HINT-AD distinguished ADNI-transferred \textit{e}MCI and \textit{l}MCI labels at the time of the speech assessment. Its measured performance therefore describes cross-sectional staging. Its contribution is the link between a stage prediction and evidence a researcher can inspect: a marker retrieves an acoustic representation or transcript span, and later stages retain that evidence when interpreting the narrative. On the same real holdout, mean ROC-AUC reached 0.77 in each language. The strongest speech baselines reached 0.67 in English and 0.73 in Mandarin. Removing phonology had the largest effect on ROC-AUC in English, whereas removing semantics had the largest effect in Mandarin. These ablations identify useful marker groups. A matched experiment that varies only the stage order would isolate the contribution of that order.

The target matters for interpretation. Age, sex, and MMSE defined the label transfer. A model using those inputs achieved ROC-AUCs of 0.91 and 0.83 (Table~\ref{tab:metadata_results}). HINT-AD recovers information about those labels from speech alone. The Mandarin interval is wide because its holdout contains nine participants (Table~\ref{tab:participant_uncertainty}). Independent same-visit staging would test clinical agreement. The next question is prognostic: can baseline speech predict progression from \textit{e}MCI to \textit{l}MCI or the onset of Alzheimer's dementia? Answering it requires longitudinal clinical outcomes over a defined follow-up period before considering use in early-intervention studies.
\label{main:end}

\clearpage
\section*{Ethics Statement}
This study analyzes existing speech datasets under their access and data-use conditions; no new participants were recruited. Controlled-access recordings remain within their repository data-use conditions. HINT-AD is evaluated as a research framework for within-MCI staging using labels transferred from ADNI reference subsets as described in Section~\ref{sec:data}. Evaluation is reported separately for English and Chinese; Spanish passage readings are used only for auxiliary training. ADNI-derived recordings are synthetic training examples, not recordings of ADNI participants speaking Mandarin.

\section*{Reproducibility Statement}
Sections~\ref{sec:method}--\ref{sec:experiments} specify the architecture, marker extraction, cohort composition, participant-disjoint split, and evaluation protocol. Appendices~\ref{appendix:hierarchical_layers} and \ref{appendix:prompting_strategy} provide marker definitions and the structured extraction prompt; Appendices~\ref{appendix:training_strategy} and \ref{appendix:computing_infra} provide the training algorithm, hyperparameters, and computing environment. All speech models use the same held-out partition and group-aware validation design. Tables~\ref{tab:performance_results}--\ref{tab:ablation_results} report performance variation across the five fold-trained models.

\section*{AI Use Statement}
AI assistants supported literature search, code implementation, interpretation of results, drafting, and language revision. The authors are responsible for the content of the submitted manuscript.

\section*{ACKNOWLEDGMENT}
We thank Professor James Rowe of the University of Cambridge for his expert validation of the speech markers used in this study.
This work was supported in part by the Technology Start-up Support Scheme for Universities (TSSSU).
This work was also supported in part by a grant from the Emergency Fund, Croucher Foundation, Hong Kong, for the project ``A cross-linguistic, family-genetic study of language in autism and the broad autism phenotype.''

\section*{CONFLICT OF INTEREST}
A patent application related to this work has been filed with the United States Patent and Trademark Office.

\bibliographystyle{template/iclr2027_conference}
\bibliography{references}
\clearpage
\appendix
\section*{Appendix: Supplementary Material}


\section{Technical Overview and Working Mechanism}
\label{appendix:technical_overview}
\begin{figure}[!ht]
  \centering
  \includegraphics[width=\linewidth]{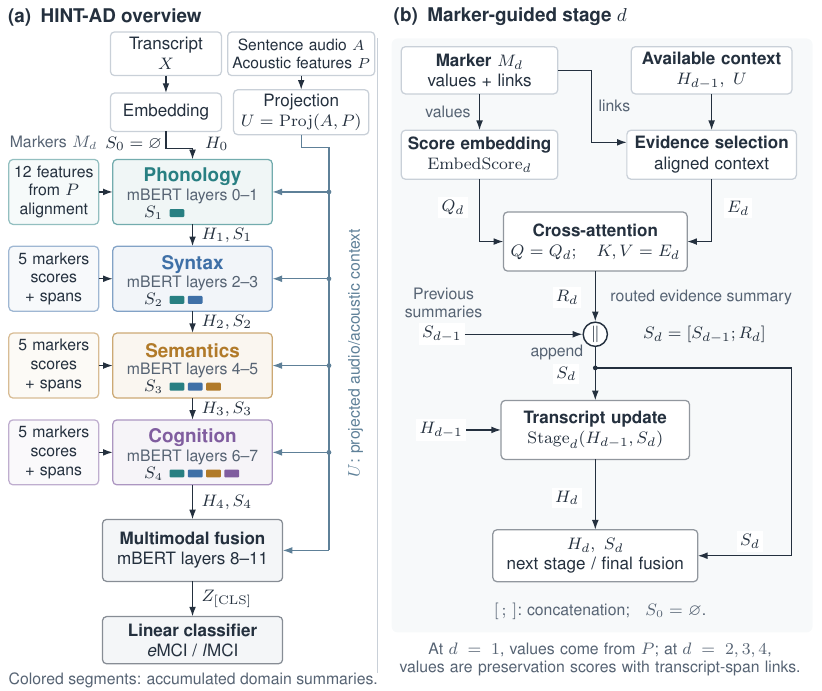}
  \caption[HINT-AD architecture and marker-guided evidence routing.]{\textbf{HINT-AD architecture and marker-guided evidence routing.}
  (a) Transcript tokens $X$ are embedded as $H_0$, and sentence-level audio
  embeddings $A$ and acoustic features $P$ are projected into $U$.
  Four linguistic stages pass updated transcript context $H_d$ and accumulated
  marker summaries $S_d$ forward. Final fusion combines $H_4$, $U$, and $S_4$
  before classification from the transcript \texttt{[CLS]} representation.
  (b) Marker values form queries $Q_d$; marker links select evidence $E_d$ from
  the available context. Cross-attention yields $R_d$, which is concatenated with
  $S_{d-1}$ before the transcript update, following Algorithm~\ref{alg:hint_training}. Phonology uses
  measured acoustic features; higher levels use language-specific preservation
  scores and transcript spans. The hierarchy is shared across languages.
  Colored segments denote domain-summary groups, not individual tokens.
  Layer indices are zero-based.}
  \label{fig:hint_ad_technical_overview}
\end{figure}

Figure~\ref{fig:hint_ad_technical_overview} complements Figure~\ref{fig:architecture} with the full architecture and an expanded marker-guided stage. It shows how marker-specific queries retrieve evidence, how summaries accumulate across linguistic levels, and how final fusion combines these summaries with transcript and audio representations.

\FloatBarrier

\section{Speech Marker Definitions and Development}
\label{appendix:hierarchical_layers}
We adapted the linguistic-marker categories of \citet{mo2024leveraging}, originally developed for AD-versus-normal-control classification, into descriptive marker names. For example, ``narrative inconsistencies'' became ``narrative sequencing.'' Using these categories as a reference, GPT-4o \citep{openai2024gpt4o} summarized syntactic, semantic, and cognitive markers from English and Chinese \textit{e}MCI and \textit{l}MCI training transcripts. The resulting marker sets were then scored using the extraction models specified in Appendix~\ref{appendix:prompting_strategy}.

The marker-development prompt was: \textit{Syntax/Grammar: e.g., structure, tense, conjunction, object vs. subject, pronoun. Semantics: e.g., meaning, synonyms, antonyms, roles distinctions (e.g., mother vs. sister). Cognition: e.g., memory loss, omissions, confusions (she vs. he), repetitions. Develop speech marker categories to guide syntax/semantics/cognition tokenization layers (with reference to the table for AD vs. NC: [AD vs. NC table]. Categorize speech markers observed in [language] transcripts of picture description tasks (e.g., Cookie Theft) into distinct categories to differentiate early mild cognitive impairment (\textup{\textit{e}MCI}) from late MCI (\textup{\textit{l}MCI}). Output a table for each level (Syntax/Grammar, Semantics, Cognition).} We also prompted GPT-4o to suggest acoustic features corresponding to the identified categories. These suggestions informed the phonological marker set; the feature values were computed from aligned recordings.

Tables~\ref{tab:phonological_markers} and \ref{tab:syntactic_semantic_cognitive} define the acoustic features and higher-level marker sets. English and Chinese share 12 phonological features and each use five markers per higher-level domain. Higher-level scores measure preservation of function; acoustic values retain their measurement interpretation.

\begin{table*}[t]
\centering
\caption{Phonological features shared by English and Chinese. Values describe acoustic measurements, not language-preservation scores.}
\label{tab:phonological_markers}
\small
\setlength{\tabcolsep}{4pt}
\begin{tabular}{@{}p{0.30\textwidth}p{0.66\textwidth}@{}}
\toprule
\textbf{Name} & \textbf{Description} \\
\midrule
Speech rate & Words per second, excluding pause segments \\
Mean pause duration within clause & Average pause duration before function words (e.g., pronouns, prepositions) \\
Clause-final pause duration & Average pause duration following clause boundaries (e.g., punctuation, conjunctions) \\
Disfluency count & Total count of filler words (e.g., ``uh'', ``um'') \\
Mean pause before content words & Average pause duration preceding content words (nouns, verbs, adjectives, adverbs) \\
Emphasis on content words & Mean pitch and energy of content words minus the global mean \\
Lexical retrieval variability & Standard deviation of pause durations before content words \\
Pause ratio & Proportion of total speaking time spent in pauses \\
Pitch variability & Standard deviation of pitch across words \\
Energy variability & Standard deviation of energy across words \\
Duration variability & Standard deviation of word durations \\
Repeated-word count & Count of locally repeated words within a sentence \\
\bottomrule
\end{tabular}
\end{table*}

\begin{table*}[t]
\centering
\caption{Syntactic, semantic, and cognitive marker sets for English and Chinese. Each assessable marker receives a preservation score from 0 to 1.}
\label{tab:syntactic_semantic_cognitive}
\small
\setlength{\tabcolsep}{3pt}
\begin{tabular}{@{}p{0.10\textwidth}p{0.41\textwidth}p{0.41\textwidth}@{}}
\toprule
\textbf{Domain} & \textbf{English} & \textbf{Chinese} \\
\midrule
Syntax &
Sentence structure; tense use; conjunction use; pronoun use; article and preposition use
&
Sentence structure; tense or time reference; conjunction use; pronoun use; word order
\\
\midrule
Semantics &
Lexical use; role identification; synonym use; contrast use; scene integration
&
Lexical use; role identification; synonym use; verb use; thematic coherence
\\
\midrule
Cognition &
Repetition; omission; narrative sequencing; referential use; story structure
&
Repetition; omission; narrative sequencing; referential use; confusion
\\
\bottomrule
\end{tabular}
\end{table*}
\FloatBarrier

\section{Illustrative Speech Examples}
\label{appendix:speech_examples}
\begin{figure}[!ht]
  \centering
  \includegraphics[width=\textwidth]{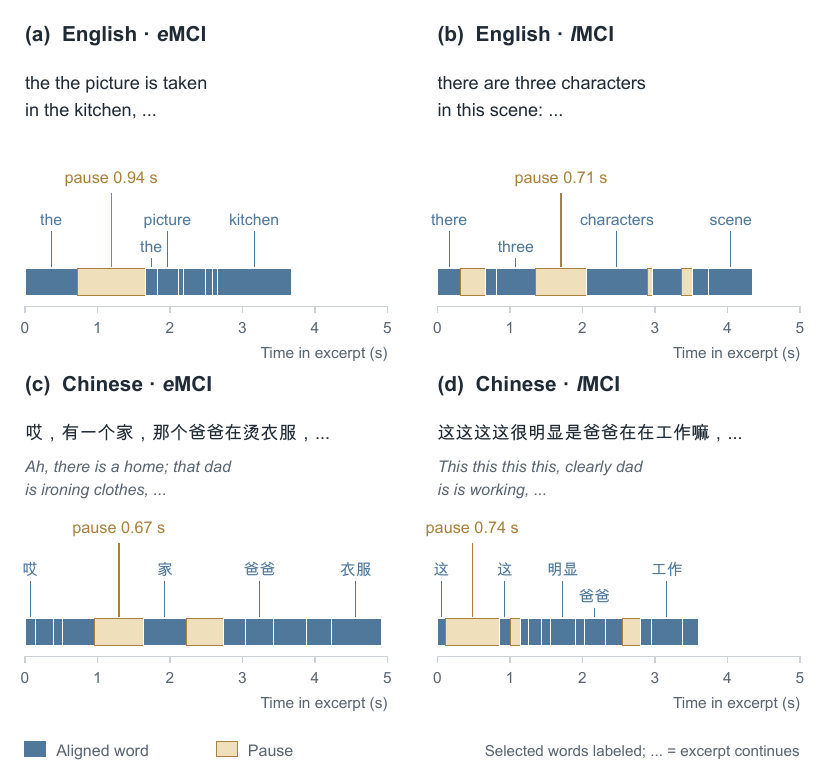}
  \caption{\textbf{Illustrative speech excerpts and forced alignment.}
  Short excerpts from four different participants cover English and Chinese
  \textit{e}MCI and \textit{l}MCI. Blue blocks show aligned word intervals;
  shaded blocks show pauses. Selected words and one pause duration per panel
  are labeled for readability. All panels use the same time scale, with time
  reset to the first displayed word; ellipses indicate that the original
  utterance continues. Chinese glosses are provided only as reading aids.
  These paired text and timing observations illustrate the inputs used to
  construct HINT-AD's sentence-level representations and acoustic features.
  Participant identifiers and recording metadata are omitted.
  The excerpts illustrate input alignment and are not intended to establish
  characteristic differences between stages.}
  \label{fig:speech_examples}
\end{figure}

Figure~\ref{fig:speech_examples} shows short excerpts of the aligned speech inputs described in Section~\ref{sec:preprocessing}.

\FloatBarrier

\section{Structured Linguistic-Marker Extraction}
\label{appendix:prompting_strategy}
For English and Chinese, each extraction call assesses one higher-level domain using the full transcript. This preserves narrative context while applying domain-specific marker definitions. The Spanish passage-reading prompt assesses only nonfunctional repetition; other higher-level markers remain null. DeepSeek-V4.1-Flash supplies the primary scores, and GPT-5.6-Luna is used for extractor-sensitivity analysis.

\textbf{Transcript Presentation.}
Each transcript is presented as a numbered list of sentences. Numbering supports direct links between a marker score and its supporting transcript spans.

\textbf{Marker Definitions.}
For each language, the marker definitions specify the phenomenon to assess and the interpretation of preserved performance. The LLM returns one transcript-level score per marker from $\{0, 0.25, 0.5, 0.75, 1\}$, where larger values indicate better-preserved language function. It may attach up to five supporting spans, each with a line number, quoted text, and local score. A marker that cannot be assessed is returned with an explicit flag and no score or evidence.

\begin{figure}[ht]
\centering
\scriptsize
\fbox{%
\begin{minipage}{0.96\columnwidth}
\ttfamily\raggedright
System:\\
You are a careful linguistic annotator of connected speech. Return a JSON response in the requested format.\\[0.4em]
User:\\
Score the requested \{domain\} markers using the full transcript. Higher scores indicate better-preserved language function.\\[0.3em]
Transcript:\\
\{numbered\_lines\}\\[0.3em]
Return only valid JSON. Do not include markdown, comments, explanations, or extra keys.\\[0.3em]
Required response format:\\
- Return exactly one object for every marker in \{marker\_set\}.\\
- Each object contains: marker, assessable, score, evidence.\\
- score must be one of 0, 0.25, 0.5, 0.75, or 1 when assessable.\\
- evidence contains at most five items with line\_number, text, and local\_score.\\
- If assessable is false, score is null and evidence is empty.\\
- Evidence text must be copied from the specified transcript line.\\[0.3em]
Return the full JSON object now.
\end{minipage}
}
\caption{Condensed prompt template for score-based marker extraction. The numbered transcript, linguistic domain, and language-specific marker set are supplied for each analysis.}
\label{fig:llm_template}
\end{figure}

\textbf{Prompt and Response Format.}
The model returns only the structured JSON response specified in Figure~\ref{fig:llm_template}. We check completeness and validity, regenerate nonconforming responses, and retain accepted annotations for reuse. Three runs are combined by within-model consensus.
\FloatBarrier

\section{Training Algorithm}
\label{appendix:training_strategy}
Algorithm~\ref{alg:hint_training} summarizes supervised training. Mini-batches contain transcript tokens $X$, sentence-level audio embeddings $A$, acoustic features $P$, marker measurements and evidence $M$, and stage labels $y$. Four linguistic stages accumulate marker summaries before final fusion with transcript and audio representations. Checkpoint selection follows Section~\ref{sec:evaluation}.

\begin{algorithm*}[ht]
\caption{HINT-AD Model Training}
\label{alg:hint_training}
\begin{algorithmic}[1]
\REQUIRE Transcripts $X$; sentence audio $A$; acoustic features $P$; marker scores and evidence $M$; labels $y$
\REQUIRE Model weights $\theta$; learning rate $\alpha$; number of epochs $N$; batch size $K$
\ENSURE Fitted model parameters $\theta$

\STATE Initialize model parameters $\theta$
\FOR{$\text{epoch} = 1$ to $N$}
    \FOR{each mini-batch $(X,A,P,M,y)$}
        \STATE \textbf{// Forward pass}
        \STATE Encode transcript tokens: $H \leftarrow \mathrm{Embed}(X)$
        \STATE Project multimodal context: $U \leftarrow \mathrm{Proj}(A,P)$; initialize summaries $S \leftarrow \emptyset$
        \FOR{$d \in \{\mathrm{phon},\mathrm{syn},\mathrm{sem},\mathrm{cog}\}$}
            \STATE Create marker queries $Q_d \leftarrow \mathrm{EmbedScore}(M_d)$
            \STATE Select marker-aligned evidence $E_d \leftarrow \mathrm{Evidence}(H,U,M_d)$
            \STATE Route evidence $R_d \leftarrow \mathrm{CrossAttn}(Q_d,E_d)$ and append $R_d$ to $S$
            \STATE Update transcript context $H \leftarrow \mathrm{Stage}_d(H,S)$
        \ENDFOR
        \STATE Fuse modalities: $Z \leftarrow \mathrm{Fusion}(H,U,S)$
        \STATE Compute predictions: $\tilde{y} \leftarrow \mathrm{Classifier}(Z_{\texttt{[CLS]}})$
        \STATE \textbf{// Loss and backpropagation}
        \STATE Compute loss: $\mathcal{L} = \operatorname{WCE}(\tilde{y}, y; w)$, where $\operatorname{WCE}$ denotes class-weighted cross-entropy with class weights $w$.
        \STATE Update parameters $\theta$ using AdamW with learning rate $\alpha$
    \ENDFOR
\ENDFOR
\STATE \textbf{return} fitted parameters $\theta$; select a checkpoint using validation ROC-AUC
\end{algorithmic}
\end{algorithm*}
\FloatBarrier

\section{Training Details and Computing Infrastructure}
\label{appendix:computing_infra}
All speech models use the same participant-disjoint train--test split and group-aware validation design.

\textbf{ADNI-derived training speech.} The 159 profiles comprised 59 \textit{e}MCI and 100 \textit{l}MCI cases staged by education-adjusted Logical Memory criteria. Eight indicators from ADNI4 English story recall guided generation: word, noun and verb counts, unique-word ratio, duration, speech rate, pause count, and mean pause length. For ADNI3, we took medians from up to 15 ADNI4 donors matched on education, MMSE and delayed recall for lexical targets; timing also used immediate recall, animal fluency and uncued picture naming. We mapped each indicator's English percentile to a range from the assigned Mandarin training group. GPT-6-Luna generated a Mandarin Cookie Theft description from these targets and examples from the assigned group, without access to the stage label or clinical scores. CosyVoice 3 synthesized the script in zero-shot mode using a reference voice from the same group.\footnote{\url{https://huggingface.co/FunAudioLLM/Fun-CosyVoice3-0.5B-2512}} The ADNI stage was the training target. Real and synthetic recordings followed the same speech preparation pipeline.
HINT-AD uses AdamW with a learning rate of $2\times10^{-5}$, batch size 8, and at most 10 epochs. Training uses class-weighted cross-entropy, a linear schedule with 10\% warmup, zero weight decay, and early stopping with patience 3. Joint checkpoints maximize mean English--Chinese validation ROC-AUC with mean AP as a tie-breaker; language-specific checkpoints use the corresponding metrics. SpeechFormer++ and HAFFormer use learning rates of $1\times10^{-4}$ and $2\times10^{-3}$, respectively, following their original settings \citep{chen2023speechformer++,dong2024hafformer}.

Training used an NVIDIA A100 40 GB GPU on a Linux system through Google Colab. The reported software environment comprises Python 3.12, PyTorch 2.12.0, and Transformers 5.13.0.

DeepSeek-V4.1-Flash was the primary model for LLM-based marker extraction. The GPT-5.6-Luna sensitivity analysis used the same prompts, response checks, retention of accepted annotations, and three-run consensus procedure.

HINT-AD used uncased multilingual BERT,\footnote{\url{https://huggingface.co/google-bert/bert-base-multilingual-uncased}} while the flat text baselines used its cased variant.\footnote{\url{https://huggingface.co/google-bert/bert-base-multilingual-cased}} All audio embeddings except those used by CogniVoice were obtained from Qwen2-Audio \citep{qwen2audio}.\footnote{\url{https://huggingface.co/Qwen/Qwen2-Audio-7B}} English, Chinese, and Spanish transcription used Qwen3-ASR-1.7B \citep{shi2026qwen3asr}.\footnote{\url{https://huggingface.co/Qwen/Qwen3-ASR-1.7B-hf}}

\begin{table}[!htbp]
\centering
\caption{Primary HINT-AD performance on held-out real recordings. Values are means across five model fits, with 95\% CIs in brackets. For each language, we resampled test participants within stage 5,000 times, keeping each participant's recordings together.}
\label{tab:participant_uncertainty}
\footnotesize
\setlength{\tabcolsep}{2pt}
\begin{tabular*}{\linewidth}{@{\extracolsep{\fill}}lcccc@{}}
\toprule
\textbf{Language} & \textbf{Test participants} & \textbf{ROC-AUC} & \textbf{\textit{e}MCI AP} & \textbf{Macro F1} \\
\midrule
English & 34 & 0.77 [0.65, 0.87] & 0.65 [0.48, 0.81] & 0.62 [0.51, 0.72] \\
Chinese & 9 & 0.77 [0.53, 0.97] & 0.67 [0.46, 0.94] & 0.62 [0.38, 0.75] \\
\bottomrule
\end{tabular*}
\end{table}

\FloatBarrier

\section{Deterministic Marker Consistency}
\label{appendix:marker_consistency}
We compare consensus repetition and omission scores with deterministic transcript measures across the 420 English and Chinese picture-description recordings. Repetition is the fraction of lexical tokens repeated within the preceding three tokens of a sentence. Content coverage is the fraction of task-specific lexical content units found in a transcript for supported picture-description tasks. Because higher marker scores indicate better-preserved function, repetition scores should correlate negatively with repeated-token frequency and omission scores positively with content coverage. Table~\ref{tab:marker_consistency} reports Spearman correlations by language and extractor.

\begin{table}[!htbp]
\centering
\caption{Spearman correlations between marker scores and transcript-derived repetition or content coverage.}
\label{tab:marker_consistency}
\setlength{\tabcolsep}{10pt}
\begin{tabular}{@{}llcc@{}}
\toprule
\textbf{Language} & \textbf{Marker} & \textbf{DeepSeek} & \textbf{GPT} \\
\midrule
English & Repetition & $-0.43$ & $-0.36$ \\
English & Omission & $+0.57$ & $+0.54$ \\
\addlinespace[2pt]
Chinese & Repetition & $-0.54$ & $-0.47$ \\
Chinese & Omission & $+0.68$ & $+0.65$ \\
\bottomrule
\end{tabular}
\end{table}

Both extractors show the expected directions in both languages. The omission association remains positive after accounting for task and transcript length (DeepSeek: 0.37 in English and 0.62 in Chinese; GPT: 0.34 and 0.61). These associations support the consistency of repetition and omission scores with independently computed transcript measures across extractors and languages.
\FloatBarrier

\section{Sensitivity to the Marker-Extraction Model}
\label{appendix:cross_llm}
Table~\ref{tab:cross_llm_results} holds the architecture, training data and real holdout fixed and changes only the source of the higher-level markers. DeepSeek markers give higher mean ROC-AUC and AP in both languages, while the higher macro F1 depends on language.

\begin{table}[!ht]
\centering
\caption{Sensitivity to marker-extraction model choice (0--1 scale). Both marker sources are evaluated with the full, ADNI-augmented training configuration. Values are mean $\pm$ SD across five fold-trained models.}
\label{tab:cross_llm_results}
\footnotesize
\setlength{\tabcolsep}{2pt}
\begin{tabular*}{\linewidth}{@{\extracolsep{\fill}}lcccccc@{}}
\toprule
 & \multicolumn{3}{c}{\textbf{English}} & \multicolumn{3}{c}{\textbf{Chinese}} \\
\cmidrule(lr){2-4}\cmidrule(l){5-7}
\textbf{Marker model} & \textbf{ROC-AUC} & \textbf{AP} & \textbf{Macro F1} & \textbf{ROC-AUC} & \textbf{AP} & \textbf{Macro F1} \\
\midrule
DeepSeek-V4.1-Flash & 0.77\,\ensuremath{\pm}\,0.03 & 0.65\,\ensuremath{\pm}\,0.04 & 0.62\,\ensuremath{\pm}\,0.03 & 0.77\,\ensuremath{\pm}\,0.10 & 0.67\,\ensuremath{\pm}\,0.11 & 0.62\,\ensuremath{\pm}\,0.10 \\
GPT-5.6-Luna & 0.73\,\ensuremath{\pm}\,0.05 & 0.62\,\ensuremath{\pm}\,0.06 & 0.65\,\ensuremath{\pm}\,0.04 & 0.68\,\ensuremath{\pm}\,0.08 & 0.54\,\ensuremath{\pm}\,0.09 & 0.55\,\ensuremath{\pm}\,0.12 \\
\bottomrule
\end{tabular*}
\end{table}

\section{Metadata baselines}
\label{appendix:metadata_baselines}
To quantify how well the transfer inputs recover the assigned stages, we rank recordings by measured MMSE alone and fit a logistic regression using standardized age, sex, and MMSE. The regression is fit on the training recordings for each language and evaluated on the same participant-disjoint holdout as the speech models. Higher MMSE scores predict \textit{e}MCI in the single-variable baseline. For each language, we select the cutoff that maximizes training macro F1 using measured MMSE scores or out-of-fold logistic regression probabilities.

\begin{table}[!ht]
\centering
\caption{Recovery of ADNI-transferred stages from the variables used for label transfer. Each value is a held-out score for one metadata baseline.}
\label{tab:metadata_results}
\small
\begin{tabular}{@{}lcccccc@{}}
\toprule
 & \multicolumn{3}{c}{\textbf{English}} & \multicolumn{3}{c}{\textbf{Chinese}} \\
\cmidrule(lr){2-4}\cmidrule(l){5-7}
\textbf{Inputs} & \textbf{ROC-AUC} & \textbf{AP} & \textbf{Macro F1} & \textbf{ROC-AUC} & \textbf{AP} & \textbf{Macro F1} \\
\midrule
MMSE alone & 0.85 & 0.74 & 0.72 & 0.75 & 0.67 & 0.68 \\
Age, sex, and MMSE & 0.91 & 0.89 & 0.84 & 0.83 & 0.64 & 0.40 \\
\bottomrule
\end{tabular}
\end{table}

\clearpage
\section{Dataset-wise results}
\label{appendix:dataset_results}
HINT-AD had the highest reported mean ROC-AUC, AP and macro F1 in both TAUKADIAL subsets (Table~\ref{tab:cohort_results}). On PROCESS-2, flat fusion led in ROC-AUC (0.66), while HINT-AD led in AP (0.52) and macro F1 (0.53). On Pitt, CogniVoice and HINT-AD had the same displayed ROC-AUC (0.85); CogniVoice had the highest AP (0.85) and macro F1 (0.62). Only five recordings contributed to the Pitt comparison.

\begin{table}[!ht]
\centering
\caption{Held-out performance by dataset. $N$ is the number of test recordings for each dataset and language. Values are mean $\pm$ SD across the same five fold-trained models as in Table~\ref{tab:performance_results}. Bold marks the highest displayed mean and underlining marks the second-highest in each column, including ties. Interpret the Pitt scores cautiously given its five test recordings.}
\label{tab:cohort_results}
\footnotesize
\setlength{\tabcolsep}{2pt}
\begin{tabular*}{\linewidth}{@{\extracolsep{\fill}}lcccccc@{}}
\toprule
 & \multicolumn{3}{c}{\textbf{Pitt ($N=5$), English}} & \multicolumn{3}{c}{\textbf{PROCESS-2 ($N=22$), English}} \\
\cmidrule(lr){2-4}\cmidrule(l){5-7}
\textbf{Model} & \textbf{ROC-AUC} & \textbf{AP} & \textbf{Macro F1} & \textbf{ROC-AUC} & \textbf{AP} & \textbf{Macro F1} \\
\midrule
Qwen2-Audio & 0.65\,\ensuremath{\pm}\,0.37 & 0.61\,\ensuremath{\pm}\,0.33 & 0.56\,\ensuremath{\pm}\,0.22 & 0.52\,\ensuremath{\pm}\,0.11 & 0.44\,\ensuremath{\pm}\,0.13 & 0.51\,\ensuremath{\pm}\,0.11 \\
mBERT & 0.65\,\ensuremath{\pm}\,0.37 & 0.61\,\ensuremath{\pm}\,0.33 & 0.33\,\ensuremath{\pm}\,0.10 & 0.43\,\ensuremath{\pm}\,0.06 & 0.33\,\ensuremath{\pm}\,0.01 & 0.41\,\ensuremath{\pm}\,0.11 \\
Flat fusion & \underline{0.75}\,\ensuremath{\pm}\,0.16 & 0.57\,\ensuremath{\pm}\,0.23 & 0.48\,\ensuremath{\pm}\,0.20 & \textbf{0.66\,\ensuremath{\pm}\,0.05} & \underline{0.48}\,\ensuremath{\pm}\,0.08 & \underline{0.52}\,\ensuremath{\pm}\,0.12 \\
Marker MLP & 0.55\,\ensuremath{\pm}\,0.29 & 0.41\,\ensuremath{\pm}\,0.12 & 0.42\,\ensuremath{\pm}\,0.03 & 0.54\,\ensuremath{\pm}\,0.03 & 0.36\,\ensuremath{\pm}\,0.01 & 0.43\,\ensuremath{\pm}\,0.05 \\
SpeechFormer++ & 0.65\,\ensuremath{\pm}\,0.25 & 0.52\,\ensuremath{\pm}\,0.26 & 0.33\,\ensuremath{\pm}\,0.14 & 0.55\,\ensuremath{\pm}\,0.09 & 0.44\,\ensuremath{\pm}\,0.12 & 0.38\,\ensuremath{\pm}\,0.14 \\
HAFFormer & 0.65\,\ensuremath{\pm}\,0.12 & 0.43\,\ensuremath{\pm}\,0.08 & 0.49\,\ensuremath{\pm}\,0.14 & 0.53\,\ensuremath{\pm}\,0.04 & 0.40\,\ensuremath{\pm}\,0.06 & 0.43\,\ensuremath{\pm}\,0.04 \\
CogniVoice & \textbf{0.85\,\ensuremath{\pm}\,0.30} & \textbf{0.85\,\ensuremath{\pm}\,0.30} & \textbf{0.62\,\ensuremath{\pm}\,0.23} & 0.61\,\ensuremath{\pm}\,0.10 & 0.46\,\ensuremath{\pm}\,0.13 & 0.50\,\ensuremath{\pm}\,0.12 \\
HINT-AD & \textbf{0.85\,\ensuremath{\pm}\,0.20} & \underline{0.77}\,\ensuremath{\pm}\,0.29 & \underline{0.61}\,\ensuremath{\pm}\,0.20 & \underline{0.63}\,\ensuremath{\pm}\,0.04 & \textbf{0.52\,\ensuremath{\pm}\,0.07} & \textbf{0.53\,\ensuremath{\pm}\,0.03} \\
\midrule
 & \multicolumn{3}{c}{\textbf{TAUKADIAL ($N=30$), English}} & \multicolumn{3}{c}{\textbf{TAUKADIAL ($N=27$), Chinese}} \\
\cmidrule(lr){2-4}\cmidrule(l){5-7}
\textbf{Model} & \textbf{ROC-AUC} & \textbf{AP} & \textbf{Macro F1} & \textbf{ROC-AUC} & \textbf{AP} & \textbf{Macro F1} \\
\midrule
Qwen2-Audio & \underline{0.70}\,\ensuremath{\pm}\,0.15 & \underline{0.67}\,\ensuremath{\pm}\,0.13 & 0.52\,\ensuremath{\pm}\,0.13 & 0.63\,\ensuremath{\pm}\,0.10 & 0.50\,\ensuremath{\pm}\,0.08 & 0.44\,\ensuremath{\pm}\,0.09 \\
mBERT & 0.67\,\ensuremath{\pm}\,0.12 & 0.63\,\ensuremath{\pm}\,0.10 & 0.47\,\ensuremath{\pm}\,0.14 & 0.65\,\ensuremath{\pm}\,0.07 & 0.59\,\ensuremath{\pm}\,0.09 & \underline{0.57}\,\ensuremath{\pm}\,0.12 \\
Flat fusion & 0.67\,\ensuremath{\pm}\,0.05 & 0.61\,\ensuremath{\pm}\,0.07 & 0.51\,\ensuremath{\pm}\,0.12 & 0.70\,\ensuremath{\pm}\,0.05 & 0.65\,\ensuremath{\pm}\,0.05 & 0.53\,\ensuremath{\pm}\,0.09 \\
Marker MLP & 0.68\,\ensuremath{\pm}\,0.13 & 0.58\,\ensuremath{\pm}\,0.08 & 0.45\,\ensuremath{\pm}\,0.13 & 0.72\,\ensuremath{\pm}\,0.14 & \underline{0.66}\,\ensuremath{\pm}\,0.18 & 0.47\,\ensuremath{\pm}\,0.15 \\
SpeechFormer++ & 0.62\,\ensuremath{\pm}\,0.12 & 0.50\,\ensuremath{\pm}\,0.08 & 0.38\,\ensuremath{\pm}\,0.12 & 0.62\,\ensuremath{\pm}\,0.13 & 0.58\,\ensuremath{\pm}\,0.13 & 0.39\,\ensuremath{\pm}\,0.14 \\
HAFFormer & 0.63\,\ensuremath{\pm}\,0.11 & 0.64\,\ensuremath{\pm}\,0.10 & 0.40\,\ensuremath{\pm}\,0.04 & \underline{0.73}\,\ensuremath{\pm}\,0.07 & 0.65\,\ensuremath{\pm}\,0.08 & 0.43\,\ensuremath{\pm}\,0.12 \\
CogniVoice & 0.68\,\ensuremath{\pm}\,0.05 & 0.63\,\ensuremath{\pm}\,0.09 & \underline{0.54}\,\ensuremath{\pm}\,0.09 & 0.65\,\ensuremath{\pm}\,0.08 & 0.50\,\ensuremath{\pm}\,0.14 & 0.46\,\ensuremath{\pm}\,0.08 \\
HINT-AD & \textbf{0.86\,\ensuremath{\pm}\,0.03} & \textbf{0.81\,\ensuremath{\pm}\,0.03} & \textbf{0.67\,\ensuremath{\pm}\,0.06} & \textbf{0.77\,\ensuremath{\pm}\,0.10} & \textbf{0.67\,\ensuremath{\pm}\,0.11} & \textbf{0.62\,\ensuremath{\pm}\,0.10} \\
\bottomrule
\end{tabular*}
\end{table}

\clearpage
\section{Stage differences in SHAP-selected markers}
\label{appendix:marker_stage}
To examine stage differences in the speech markers identified by SHAP, we selected ten markers per language: the five with the strongest contributions toward \textit{e}MCI and the five with the strongest contributions toward \textit{l}MCI (Table~\ref{tab:marker_shap}). For each marker, we averaged observed values across folds for each recording and then across recordings for each participant, excluding missing values. We used two-sided Welch's $t$-tests to compare these participant means between the two transferred stage groups, without adjustment for multiple comparisons.

\begin{table}[!ht]
\centering
\caption{SHAP-selected markers by language and direction of attribution. Stage means are on a 0--1 scale. Higher linguistic scores indicate better preservation, and acoustic measures are normalized using training data. $\Delta$ is the \textit{e}MCI mean minus the \textit{l}MCI mean. Bold indicates two-sided $p<0.05$ (Welch's $t$-test).}
\label{tab:marker_stage}
\footnotesize
\setlength{\tabcolsep}{3pt}
\begin{tabularx}{\linewidth}{@{}cl>{\raggedright\arraybackslash}Xrrrrr@{}}
\toprule
 & & & & \multicolumn{2}{c}{\textbf{Mean marker value}} & & \\
\cmidrule(lr){5-6}
\textbf{Rank} & \textbf{Level} & \textbf{Marker} & \textbf{Mean SHAP} & \textbf{\textit{e}MCI} & \textbf{\textit{l}MCI} & $\boldsymbol{\Delta}$ & $\boldsymbol{p}$ \\
\midrule
\multicolumn{8}{@{}l}{\textbf{English: toward \textit{e}MCI (positive SHAP values)}} \\
1 & Semantics & Role identification & $+0.01038$ & 0.8813 & 0.7849 & $+0.0963$ & 0.1294 \\
2 & Semantics & Lexical use & $+0.00855$ & 0.7751 & 0.7001 & $+0.0751$ & \textbf{0.0297} \\
3 & Semantics & Scene integration & $+0.00851$ & 0.8046 & 0.7123 & $+0.0923$ & \textbf{0.0197} \\
4 & Semantics & Contrast use & $+0.00416$ & 0.6302 & 0.5883 & $+0.0419$ & 0.5600 \\
5 & Semantics & Synonym use & $+0.00171$ & 0.5556 & 0.5152 & $+0.0404$ & 0.2763 \\
\midrule
\multicolumn{8}{@{}l}{\textbf{English: toward \textit{l}MCI (negative SHAP values)}} \\
1 & Phonology & Mean pause duration within clause & $-0.00069$ & 0.1880 & 0.2645 & $-0.0765$ & \textbf{0.0406} \\
2 & Phonology & Word-duration variability & $-0.00059$ & 0.5303 & 0.6101 & $-0.0798$ & \textbf{0.0141} \\
3 & Phonology & Clause-final pause duration & $-0.00052$ & 0.2282 & 0.2917 & $-0.0634$ & 0.1025 \\
4 & Phonology & Energy variability & $-0.00039$ & 0.4881 & 0.5802 & $-0.0921$ & 0.1569 \\
5 & Phonology & Pitch variability & $-0.00022$ & 0.5555 & 0.5596 & $-0.0042$ & 0.8930 \\
\midrule
\multicolumn{8}{@{}l}{\textbf{Chinese: toward \textit{e}MCI (positive SHAP values)}} \\
1 & Semantics & Thematic coherence & $+0.00228$ & 0.6875 & 0.4755 & $+0.2120$ & \textbf{0.0003} \\
2 & Semantics & Role identification & $+0.00177$ & 0.7338 & 0.5116 & $+0.2222$ & \textbf{0.0048} \\
3 & Semantics & Verb use & $+0.00169$ & 0.6273 & 0.5019 & $+0.1255$ & 0.0643 \\
4 & Semantics & Lexical use & $+0.00144$ & 0.6181 & 0.4891 & $+0.1289$ & \textbf{0.0344} \\
5 & Phonology & Energy variability & $+0.00030$ & 0.6491 & 0.7330 & $-0.0838$ & 0.5446 \\
\midrule
\multicolumn{8}{@{}l}{\textbf{Chinese: toward \textit{l}MCI (negative SHAP values)}} \\
1 & Phonology & Pause ratio & $-0.00041$ & 0.5065 & 0.5521 & $-0.0456$ & 0.5891 \\
2 & Phonology & Disfluency count & $-0.00025$ & 0.3843 & 0.3758 & $+0.0085$ & 0.9722 \\
3 & Semantics & Synonym use & $-0.00017$ & 0.4167 & 0.3414 & $+0.0752$ & 0.3013 \\
4 & Phonology & Mean pause before content words & $-0.00012$ & 0.3706 & 0.4149 & $-0.0442$ & 0.5847 \\
5 & Phonology & Clause-final pause duration & $-0.00010$ & 0.3891 & 0.3946 & $-0.0055$ & 0.9510 \\
\bottomrule
\end{tabularx}
\end{table}

Seven of 20 comparisons yielded $p<0.05$ (Table~\ref{tab:marker_stage}). Lexical use was the only selected marker meeting this threshold in both English ($p=0.0297$) and Mandarin ($p=0.0344$).

In English, the \textit{e}MCI group had higher lexical-use and scene-integration scores, shorter pauses before function words and less variable word durations. In Mandarin, all three differences were semantic: the \textit{e}MCI group scored higher on thematic coherence, role identification and lexical use.

\clearpage
Tables~\ref{tab:marker_examples_en} and~\ref{tab:marker_examples_zh} illustrate the seven comparisons with $p<0.05$ using same-task excerpt pairs. For English we report mean inter-word pauses longer than 0.2~s before function words, or the population standard deviation of aligned word durations within an excerpt, excluding punctuation and pause tokens. These examples illustrate the measures rather than representing every participant.

\begin{table}[!ht]
\centering
\caption{English picture-description excerpts for SHAP-selected markers. Bold annotations give timing measurements in seconds.}
\label{tab:marker_examples_en}
\small
\begin{tabularx}{\linewidth}{@{}>{\raggedright\arraybackslash}X@{\hspace{1em}}>{\raggedright\arraybackslash}X@{}}
\toprule
\textbf{\textit{e}MCI excerpt} & \textbf{\textit{l}MCI excerpt} \\
\midrule
\multicolumn{2}{@{}l}{\textbf{Lexical use} (semantics)} \\
``she's left the tap on, and the sink is overflowing and forming a puddle under the sink, which she's standing in.'' & ``no, he's something in his hand.'' \\
\multicolumn{2}{@{}p{\linewidth}@{}}{The \textit{e}MCI speaker names the tap and the puddle. The \textit{l}MCI speaker uses the word ``something.''} \\
\midrule
\multicolumn{2}{@{}l}{\textbf{Scene integration} (semantics)} \\
``and the mother is washing dishes, but she's very interested in looking on what's going on outside and not paying attention to the water spilling out, getting her feet wet.'' & ``there's water filling on the floor.'' \\
\multicolumn{2}{@{}p{\linewidth}@{}}{The \textit{e}MCI speaker links the mother's diverted attention to the spill and its consequences. The \textit{l}MCI speaker describes the water on the floor.} \\
\midrule
\multicolumn{2}{@{}l}{\textbf{Mean pause duration within clause} (phonology)} \\
``the \textbf{[0.31~s]} the mother is drying a plate \textbf{[0.61~s]} whilst looking over the garden \textbf{[0.24~s]} through a window.''\par\smallskip\textbf{Mean pause: 0.39~s.} & ``in the kitchen \textbf{[2.51~s]} there is a window \textbf{[0.80~s]} with grass \textbf{[0.24~s]} with garden outside.''\par\smallskip\textbf{Mean pause: 1.18~s.} \\
\multicolumn{2}{@{}p{\linewidth}@{}}{The mean pause before function words is shorter in the \textit{e}MCI excerpt (0.39~s vs.\ 1.18~s).} \\
\midrule
\multicolumn{2}{@{}l}{\textbf{Word-duration variability} (phonology)} \\
``the the mother is drying a plate whilst looking over the garden through a window.''\par\smallskip\textbf{Word-duration SD: 0.20~s.} & ``the child is falling from the chair as he's grabbing a cookie jar.''\par\smallskip\textbf{Word-duration SD: 0.47~s.} \\
\multicolumn{2}{@{}p{\linewidth}@{}}{Word durations vary less within the \textit{e}MCI excerpt (SD 0.20~s) than within the \textit{l}MCI excerpt (0.47~s).} \\
\bottomrule
\end{tabularx}
\end{table}

\clearpage
\begin{table}[!ht]
\centering
\caption{Chinese picture-description excerpts for SHAP-selected markers. Italics give English translations. Bracketed ellipses mark omissions.}
\label{tab:marker_examples_zh}
\small
\begin{tabularx}{\linewidth}{@{}>{\raggedright\arraybackslash}X@{\hspace{1em}}>{\raggedright\arraybackslash}X@{}}
\toprule
\textbf{\textit{e}MCI excerpt} & \textbf{\textit{l}MCI excerpt} \\
\midrule
\multicolumn{2}{@{}l}{\textbf{Thematic coherence} (semantics)} \\
\zh{一个爸爸在烫衣服，然后有一小孩在玩那个，要把那个插头，啊，狗要去阻止他，因为怕小孩电到。 然后冲过去的时候，猫吓到跳高，把那个花瓶的水打翻了，把那个棉被都淋湿了。}\par\smallskip\textit{A father is ironing, then a child is playing with that, about to take that plug, ah, the dog wants to stop him because it is afraid the child will get an electric shock. Then, as it rushes over, the cat jumps in fright, knocks over the water in that vase, and gets that quilt all wet.} & \zh{他去，狗要去抓小孩呀，嗯，啊，那个猫看到了，那个这个大人吓了一大跳，这样子对吗？ [...] 这边就是狗啊，跟小朋友啊，嗯，对呀，这是什么我也不知道。}\par\smallskip\textit{He goes, the dog wants to grab the child, um, ah, that cat sees it, that, this adult is startled, is that right? [...] Over here is the dog, and the child, um, yes, I do not know what this is either.} \\
\multicolumn{2}{@{}p{\linewidth}@{}}{The \textit{e}MCI speaker connects the danger, the dog's intervention, and its consequences. The \textit{l}MCI speaker mentions the figures but leaves their relationship unclear.} \\
\midrule
\multicolumn{2}{@{}l}{\textbf{Role identification} (semantics)} \\
\zh{好，就有两个老先生在下棋，有喝茶，然后有两个小朋友在打羽毛球 [...]}\par\smallskip\textit{Well, there are two older men playing chess and drinking tea, then two children playing badminton [...]} & \zh{这个打乒乓球嘛，阿做做象棋 [...]}\par\smallskip\textit{This, playing table tennis, ah do do chess [...]} \\
\multicolumn{2}{@{}p{\linewidth}@{}}{The \textit{e}MCI speaker identifies who is playing each game. The \textit{l}MCI speaker names activities in fragments.} \\
\midrule
\multicolumn{2}{@{}l}{\textbf{Lexical use} (semantics)} \\
\zh{[...] 然后有两个小朋友在打羽毛球，然后另外一个小男生在遛狗，还有两个小朋友在荡秋千 [...]}\par\smallskip\textit{[...] then two children are playing badminton, another little boy is walking a dog, and two more children are on the swings [...]} & \zh{这一会看了他跳棋、玩象棋啊，荡秋千啊，那个什么滑溜、滑滑什么盘啊？}\par\smallskip\textit{Now, looking at him playing checkers, playing chess, on the swings, that what, slippery, slide-slide what board?} \\
\multicolumn{2}{@{}p{\linewidth}@{}}{The \textit{e}MCI speaker names activities such as playing badminton and walking a dog. The \textit{l}MCI speaker searches for a word and repeats fragments.} \\
\bottomrule
\end{tabularx}
\end{table}

\end{document}